\documentclass[11pt]{article}

\usepackage[final]{acl}

\usepackage{times}
\usepackage{latexsym}
\usepackage[T1]{fontenc}
\usepackage[utf8]{inputenc}
\usepackage{amsmath}
\usepackage{microtype}
\usepackage{inconsolata}
\usepackage{graphicx}
\usepackage{enumitem}
\usepackage{booktabs}
\usepackage{tabularx}
\usepackage{placeins}
\usepackage{tikz}
\usetikzlibrary{positioning,arrows.meta}

\newcommand{\nPapers}{748}
\newcommand{\nProcessedPapers}{4,488}
\newcommand{\nClaims}{16,576}
\newcommand{\avgClaimsTwenty}{3.58}
\newcommand{\avgClaimsTwentyFive}{4.01}
\newcommand{\extractor}{qwen3-235b-a22b-thinking-2507}

\newcommand{\demopage}{https://hamyrappy.github.io/drift-inspector}
\newcommand{\repopage}{https://github.com/hamyrappy/drift-inspector}
\newcommand{\demopageEmnlp}{https://hamyrappy.github.io/drift-inspector-emnlp/}
\newcommand{\demopageScale}{https://hamyrappy.github.io/drift-inspector-acl/}
\newcommand{\demourl}{\url{\demopage}}

\newcommand{\demourlScale}{\url{\demopageScale}}

\newcommand{\repohref}[1]{\href{\repopage}{#1}}

\title{Drift Inspector: Exploring and Measuring Scientific Drift with \\ Atomic Contribution Claims}

\author{
  \textbf{Vsevolod Karimov\textsuperscript{1,2}}\textsuperscript{*},
  \textbf{Stepan Ostarkov\textsuperscript{3,4}},
  \textbf{Anastasia Poroshina\textsuperscript{1}}, \\
  \textbf{Anatoly Frolov\textsuperscript{1,5}}, and
  \textbf{Alexander Panchenko\textsuperscript{1,5}}\textsuperscript{*}
\\
  \textsuperscript{1}Skoltech,
  \textsuperscript{2}HSE University,
  \textsuperscript{3}Lomonosov Moscow State University,
  \textsuperscript{4}ITMO University, \textsuperscript{5}AIRI
}

\begin{document}
\maketitle
\renewcommand{\thefootnote}{*}
\footnotetext{\raggedright Corresponding authors:~\href{mailto:hamyrappy@gmail.com}{\{hamyrappy, panchenko.alexander\}@gmail.com}}
\renewcommand{\thefootnote}{\arabic{footnote}}

\begin{abstract}
Scientific abstracts mix contributions with background, motivation, and
meta-language, so tools that read them as-is cannot separate what a field
\emph{produces} from what it \emph{discusses}. We present \emph{Drift
Inspector}, an open-source system for measuring and exploring how a research
field changes over time at the level of \emph{Atomic Contribution Claims (ACCs)}:
decontextualized, contribution-bearing propositions an LLM extracts
from each abstract before analysis. The system clusters these
claims across years into an interactive map where every trend traces back
to the claims and papers behind it. Applied to six years
of EMNLP, it shows the field shifting away from classic NLP tasks toward
LLM-era capabilities such as reasoning and multimodality \textemdash{} a movement
that keyword or whole-abstract counts blur. The released data extend beyond EMNLP:
the same pipeline has processed the full ACL Anthology (346k claims, 80k
abstracts, 423 venues). Extraction is human-validated and
clustering checked against an external manually constructed taxonomy.\footnote{A live demo, code and data are available at:\ \demourl{}.}
\end{abstract}

\section{Introduction}

Major NLP venues, such as ACL or EMNLP, publish thousands of papers released every few months --- far beyond what anyone can read. This is due to recent acceleration in AI development, firstly attracting more interest to the field, but also because the research itself, including paper writing, now can be automated with LLMs and agents. As a result, the number of published papers has a trend on sharp increase every year. A similar situation is observed in most large AI-related conferences beyond NLP, such as NeurIPS, ICLR, CVPR, etc.

This urges development of tools for automatic summarization and effective and efficient exploration of this tsunami of research contributions. Besides, to understand how a research field evolves, one needs to know what its papers contribute, not only what they talk about. In rapidly moving fields like NLP the two diverge: an old thread may still saturate the motivation of papers that now advance other topics, while a new direction may appear only in contribution statements.

\begin{figure*}[t!]
\centering
\includegraphics[width=0.96\linewidth]{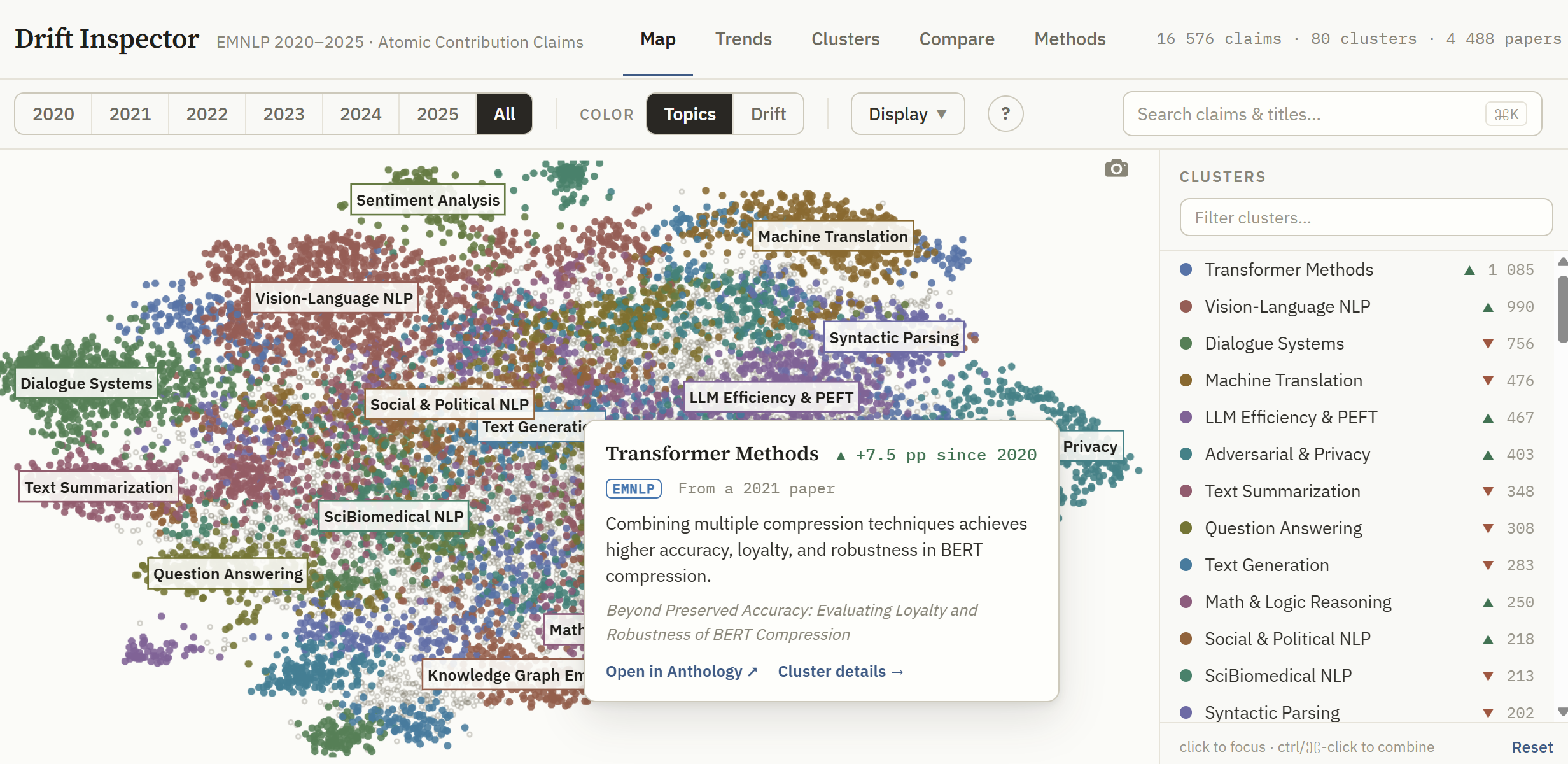}
\vspace{-2mm}
\caption{\textbf{Drift Inspector frontend} in Map view representing an interactive WebGL map of \nClaims{} atomic claims from EMNLP 2020--2025. Top bar tabs allow to switch between views (Map, Trends, Clusters, Compare or Methods). A user can adjust year filter, color mode (Cluster or Drift Trend), or perform a claim search with a per-cluster match ranking. On the right are filterable list of clusters. Hovering a point shows its claim and the cluster's prevalence trajectory; clicking pins the card with a link to the source paper.}
\label{fig:inspector}
\vspace{-2mm}
\end{figure*}

Existing literature exploration systems do not make this distinction. For instance, NLP Scholar \citep{mohammad-2020-nlp-scholar}, NLPExplorer \citep{parmar2020nlpexplorer}, and CL Scholar \citep{singh-etal-2018-cl} index papers, venues, authors, and term frequencies; topic-model dashboards built on LDA or BERTopic \citep{Grootendorst2022BERTopicNT} operate on raw abstract text. So in existing systems, usually, the word ``parsing'' would count the same whether it appears in the motivation of an LLM reasoning paper or in the actual contribution of a dependency parsing paper.

\textbf{Drift Inspector}, the system presented in this paper, aims to address this gap. Its pipeline first maps each abstract into a small set of \textbf{Atomic Contribution Claims (ACCs)} --- self-contained, falsifiable propositions about what the paper contributes, stripped of motivation, prior-work framing, and meta-language.
Extraction acts as an analytic lens: it fixes what kind of information enters every downstream statistic. We then embed the claims with SPECTER2, cluster them jointly across years, and aggregate them into per-cluster prevalence trajectories.
The result is served as a web application with linked views: an interactive claim map (year filters, drift coloring), cluster trend pages, side-by-side comparison of cohorts --- conferences, authors, keyword subsamples (Figure~\ref{fig:aclcoling}); every aggregate traces back to the claims and source papers.

The system is aimed at three audiences: (i) \emph{researchers} positioning new work or writing surveys, who need to see how a topic evolved and what replaced it; (ii) \emph{conference organizers} or \emph{area chairs}, who need evidence of topical shifts across conference years; and (iii) \emph{science-of-science researchers}, who get a validated, reusable extraction pipeline.

Our contributions are three-fold:
\begin{itemize}[noitemsep, topsep=2pt, leftmargin=15pt]
    \item A drift-measurement method based on Atomic Contribution Claim (ACC) extraction.

    \item Pre-extracted results for EMNLP and ACL Anthology corpora.
    The extracted claims were validated by human annotators and against gold reference topics.

    \item A visualization and exploration tool of the ACC datasets enabling a drift analysis across years, authors, conferences, etc. (cf. Figure~\ref{fig:inspector}).
\end{itemize}

\section{Related Systems and Work}
\label{sec:related}

\begin{figure*}[t]
\centering
\includegraphics[width=\textwidth]{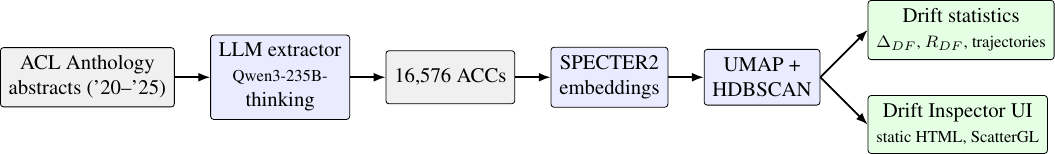}
\caption{\textbf{System workflow.} An LLM maps each abstract to Atomic Contribution Claims; claims are embedded with SPECTER2 and clustered jointly across years; per-cluster document-frequency trajectories and the interactive map are generated from the same canonical clustering.}
\label{fig:architecture}
\end{figure*}

\subsection{Literature exploration systems}

NLP Scholar \citep{mohammad-2020-nlp-scholar} provides interactive dashboards over ACL Anthology metadata and term frequencies; NLPExplorer \citep{parmar2020nlpexplorer} and CL Scholar \citep{singh-etal-2018-cl} index papers, fields, and citation structure. These systems operate at the topic or keyword level.
More recent systems include NLP-KG~\cite{schopf-matthes-2024-nlp}, GenGo~\cite{takeshita-etal-2024-gengo}, and GenGo Ultra~\cite{takeshita-etal-2025-gengo}. Various surveys of NLP as a field have also been conducted, e.g.\ \cite{schopf-etal-2023-exploring}.

Drift Inspector differs in its unit of analysis, extracting \emph{atomic contribution claims}, distinguishing what papers contribute from what they mention. As a tool it also adds a drift-color overlay that turns the claim map into a field-level trend heatmap, claim-to-paper traceability from every aggregate, side-by-side cohort comparison (venues, authors, keyword-defined subsamples), and a corpus-agnostic single-file deploy that these corpus-tied dashboards do not offer.

\subsection{Diachronic analyses of NLP}
Prior work tracks task/method prevalence \citep{uban-etal-2021-studying}, idea co-occurrence \citep{tan-etal-2017-friendships}, rhetorical framing \citep{prabhakaran-etal-2016-predicting}, and causal paradigm shifts \citep{pramanick-etal-2023-diachronic}. Closest is \citet{pramanick-etal-2025-nature}, which classifies contribution statements into a fixed taxonomy. Our system instead lets contribution clusters emerge bottom-up and packages the whole measurement loop --- extraction, validation, clustering, statistics, visualization --- into a reusable tool.

\subsection{Atomic semantic units}
Claim-level decomposition follows the atomization principle of \text{FActScore} \citep{min-etal-2023-factscore} and Atomic Content Units \citep{liu-etal-2023-revisiting}, used there for factuality and summarization evaluation. Idea-atom decompositions have also been used to sample novel research directions \citep{artiles2026alien}. We repurpose atomization for scientometric measurement. ACCs are units of field-level counting, so we check them against their own abstracts and leave their truth to the source paper.

\section{System Description}
\label{sec:system}

Figure~\ref{fig:architecture} shows an overall workflow of Drift Inspector featuring offline part (extraction, embedding, clustering, drift statistics) and the online part, which was illustrated earlier in detail in Figure~\ref{fig:inspector}.

\subsection{Atomic Contribution Claims}
\label{subsec:accs}
Raw abstracts conflate multiple contributions, background context, and meta-language (``In this paper, we propose...''). Given a document $d$, an LLM agent maps its abstract into a set of ACCs $A_d = \{a_1, \dots, a_n\}$, each satisfying three constraints: \textbf{atomicity} (exactly one contribution-bearing proposition), \textbf{decontextualization} (pronouns resolved to named entities, meta-language removed), and \textbf{falsifiability} (a verifiable assertion). A typical ACC looks like ``\emph{TheoremLlama uses curriculum learning and block training techniques to train large language models for formal theorem proving}'' \cite{2024.emnlp-main.667}. We deliberately operate on abstracts: they are the author-curated contribution summary, are uniformly available across all six years, and keep the pipeline to one cheap LLM pass per paper. The extractor is \texttt{\extractor} \citep{qwen3technicalreport} (via OpenRouter, temperature 0.2) with a few-shot prompt that excludes background, motivation, and raw metric claims; the condensed prompt is in Appendix~\ref{sec:app_extraction} and the full prompt in the released code. On EMNLP 2020--2025 the extractor yields \avgClaimsTwenty{} claims per abstract in 2020 and \avgClaimsTwentyFive{} in 2025; abstracts with no extractable contribution (3 of 751 in 2020) are excluded.

\subsection{Semantic Topology}
\label{subsec:topology}
We embed all claims with SPECTER2 \citep{cohan-etal-2020-specter,singh-etal-2023-scirepeval}, a scientific encoder pretrained on citation relatedness, so claims about similar mechanisms land close together. We then reduce the vectors with UMAP \citep{McInnes2018UMAPUM} and cluster them with HDBSCAN \citep{mcinness2017hdbscan}, as in BERTopic \citep{Grootendorst2022BERTopicNT}, except that our unit is the claim rather than the abstract. Each cluster gets class-based TF-IDF descriptors --- short (top-3, e.g., \textit{agents, action, web}) and extended (top-5) --- and all 80 clusters additionally carry LLM-generated, author-reviewed readable names: a short name for map labels (e.g., \emph{Math \& Logic Reasoning}) and a full name on the cluster page (e.g., \emph{Mathematical and Logical Reasoning in LLMs}), with the c-TF-IDF descriptor preserved in the claim card and cluster page.

About 36\% of claims remain unclustered ``noise'' --- paper-specific contributions that have not (yet) become recurring field-level patterns; the share rises from 31--34\% in 2020--2022 to 42\% in 2025, consistent with the newest contributions having had the least time to consolidate. These are kept in the interface as a background layer rather than discarded. Retaining them is why the pipeline clusters by density (HDBSCAN): a method that must place every claim would put those with no field-level counterpart into the nearest cluster and dilute it. All hyperparameters are listed in Appendix~\ref{sec:app_representation}.

\subsection{Drift Quantification}
\label{subsec:drift}
We measure prevalence as paper-level \textbf{document frequency}: the fraction of papers in year $y$ with at least one claim from cluster $c$,
$$P_y(c) = |\{d \in D_y \mid A_d \cap c \neq \emptyset\}| \, / \, |D_y|$$
where $D_y$ is the set of papers in year~$y$ and $A_d$ the claims extracted from paper~$d$.
This prevents prolific claim lists from inflating a topic: a paper with ten RAG claims counts once. Because $P_y(c)$ is computed independently per cluster, the unclustered noise does not renormalize the other clusters' shares. The system reports the absolute endpoint shift $\Delta_{DF}(c) = P_{2025}(c) - P_{2020}(c)$, the relative change $R_{DF}(c)$, full per-year trajectories, and a relative \emph{drift} score --- the base-2 log-ratio of the 2025 to 2020 share --- that drives the map's drift-color overlay (saturating at an $8\times$ change).

\subsection{The Drift Inspector Frontend}
\label{subsec:frontend}
The frontend (Figure~\ref{fig:inspector}) is a web application with five linked views. Every number it shows opens down to the claims and papers behind it.

\begin{itemize}[noitemsep, topsep=2pt, leftmargin=15pt]
    \item \textbf{Map.} A WebGL scatter (Plotly ScatterGL) of all \nClaims{} claims over a shared 2D UMAP projection. Year-filter buttons show any single year (2020--2025) or all at once; a color switch recolors every cluster by its prevalence trend (red = declining, green = growing), turning the map into a field-level drift heatmap; a Display menu toggles cluster names, the noise layer, and auto-fit. A search box highlights all claims whose text or source title contains a query substring (e.g., \texttt{retrieval}) and ranks the matching clusters. Hovering a point shows the claim, source paper, year, cluster name, and prevalence trajectory; clicking pins this card with a link to the ACL Anthology. The legend isolates clusters (ctrl/cmd-click to combine).
    \item \textbf{Trends.} A butterfly chart of the 2020$\to$2025 document-frequency shift, per-year trajectories for any clusters the user selects (by clicking bars or sparklines), and a sparkline overview of all clusters.
    \item \textbf{Clusters.} A per-cluster profile: its document frequency by year, its location in claim space, and all of its claims with links to source papers, exportable as CSV.
    \item \textbf{Compare.} Side-by-side thematic profiles of two user-chosen cohorts (a conference, an author, or a keyword-defined subsample), ranking clusters by their gap in paper share (Figure~\ref{fig:aclcoling}), with a paper/claim-share toggle, per-cohort year ranges, and an \emph{over time} mode (Figure~\ref{fig:compare}).
    \item \textbf{Methods.} A self-contained account of the pipeline, corpus, and views, generated from the corpus metadata so the build re-labels itself for a new dataset.
\end{itemize}
The noise layer renders as a faint background on the Map and can be inspected like any other point. A light/dark theme toggle and the Display options reconfigure the view for figure capture.

\subsection{Implementation and Availability}
\label{subsec:availability}
The pipeline is Python (transformers, adapters, umap-learn, hdbscan, Plotly); every stage caches its artifact (claims CSV, embedding matrix, projection, cluster assignments), and all stochastic steps are seeded, so from the released claims the published clustering is exactly reproducible from configuration. Re-running the extraction reproduces aggregate rates but not individual claim strings, because LLM decoding is not reproducible run to run. Anybody can rerun the whole process end to end for a small API cost and negligible compute (\S\ref{subsec:scale}). The frontend has no dependencies beyond a browser
and ships as a self-contained file in our \repohref{GitHub repository}. Code is released under the MIT license; the claim dataset under CC~BY~4.0 (derived from openly licensed ACL Anthology abstracts, CC~BY~4.0).
Appendix~\ref{sec:app_representation} gives the extractor and its decoding parameters, the embedding checkpoint, the clustering hyperparameters and the random seeds.

\begin{figure}[t]
\centering
\includegraphics[width=\linewidth]{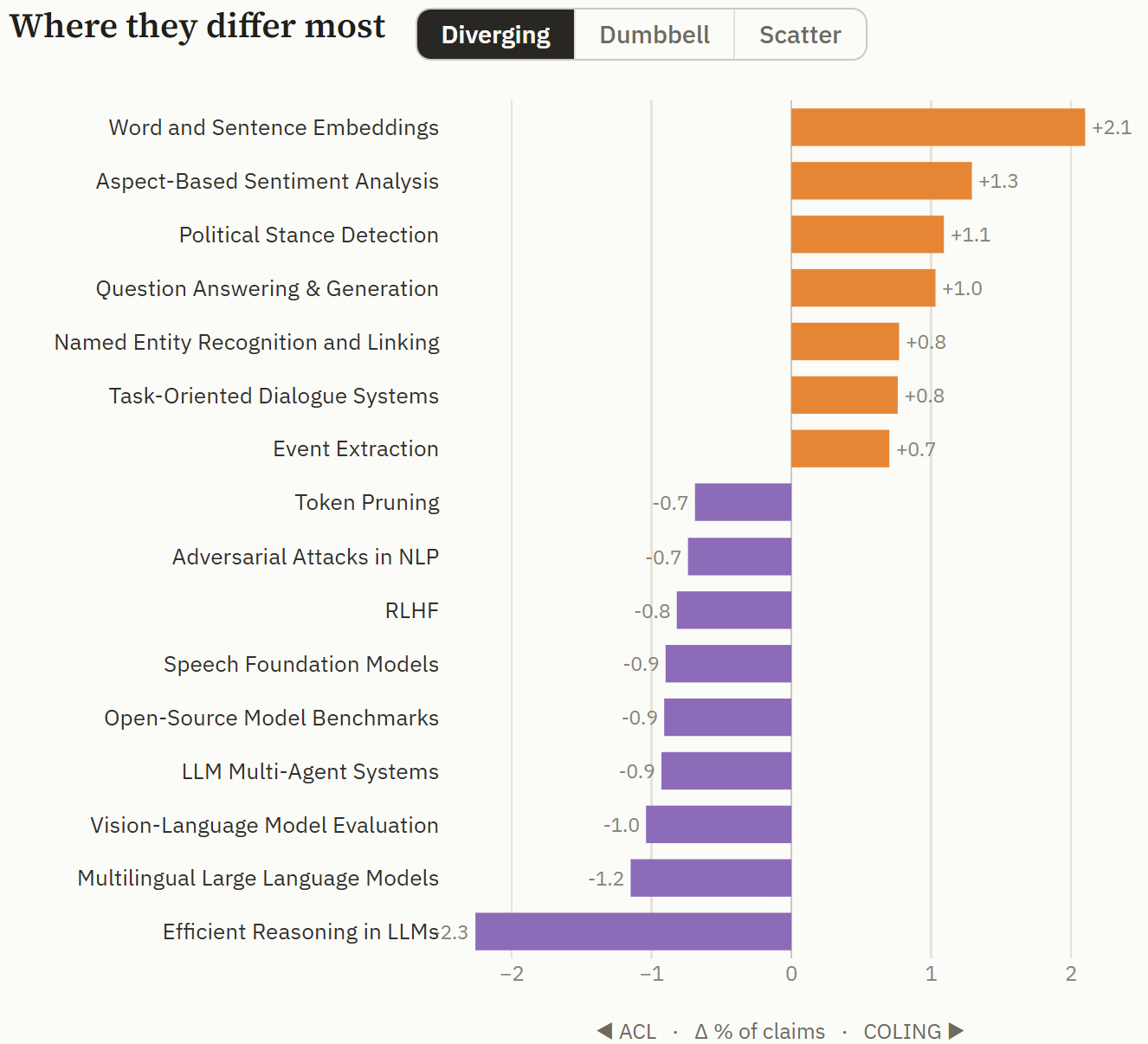}
\vspace{-4mm}
\caption{\textbf{Compare view, venue cohorts} (live six-venue instance): where ACL and COLING differ most, as the gap in cluster claim shares. COLING keeps a markedly stronger classic-NLP profile (embeddings, aspect-based sentiment, NER, event extraction), while ACL leans toward LLM-era clusters (efficient reasoning, multilingual LLMs, vision--language evaluation). Chips switch the diverging / dumbbell / scatter renderings; the same view compares authors and keyword-defined cohorts.}
\label{fig:aclcoling}
\vspace{-2mm}
\end{figure}

The same pipeline runs unchanged at corpus scale. On a research inference cluster we extracted the full ACL Anthology --- 80{,}144 abstracts, 423 venues, 346{,}010 claims --- with the cheaper \texttt{gpt-oss-120b} extractor ($\sim$\$40 at median commercial API prices; \S\ref{subsec:scale}), and we run a second live Drift Inspector over six *ACL main conferences (70k claims, 2018\textendash 2026).\footnote{\demourlScale} There, cross-venue and cross-author questions become one-click cohort comparisons: Figure~\ref{fig:aclcoling} contrasts ACL and COLING directly. On the EMNLP corpus this extractor swap preserves cluster validity and all 15 headline drift directions (\S\ref{subsec:robustness}). The human-aligned judge (\S\ref{subsec:human}) rates 300-claim samples from the scaled corpora at 96.7--99.7\% \text{Good} (full EMNLP main track 98.7\%, full ACL 99.7\%, whole anthology 96.7\%) against a 98.3\% same-judge control on the validated corpus --- overlapping 95\% intervals throughout --- and per-corpus judge reports ship with the data. All analyses in this paper are computed on the human-validated EMNLP corpus.

\subsection{Scalability and Cost}
\label{subsec:scale}
The pipeline makes one LLM call per abstract, so extraction dominates the cost and grows linearly with the corpus (Table~\ref{tab:scale}). Budgeting a new corpus is one multiplication: about \textbf{\$0.50 per thousand abstracts} with \texttt{gpt-oss-120b}, so the full ACL Anthology --- 80{,}144 abstracts --- costs $\sim$\$40 and runs overnight. A reasoning extractor such as Qwen3-235B-Thinking costs $\sim$\$11 per thousand instead, because most of what you pay for is hidden reasoning, not the claims it returns. The cheap extractor preserves every headline drift direction (\S\ref{subsec:robustness}), so you do not need to pay it. Everything after extraction is two orders of magnitude cheaper and needs no accelerator: embedding, projection and clustering finish in well under an hour of laptop CPU at either scale, and the frontend needs no server.

\begin{table}[t]
\centering
\small
\setlength{\tabcolsep}{2pt}
\begin{tabular}{l r r}
\toprule
 & \textbf{EMNLP '20--'25} & \textbf{ACL Anthology} \\
\midrule
Abstracts & 4{,}960 & 80{,}144 \\
Claims & 18{,}293 & 346{,}010 \\
Extractor & Qwen3-235B-T & gpt-oss-120b \\
Extraction cost & \$54 & $\sim$\$40 \\
\quad per 1k abstracts & \$10.87 & \$0.51 \\
Extraction wall-clock & 1.9\,h & 6.8\,h \\
Embed + cluster$^{*}$ & 8\,min & 34\,min \\
Frontend, one file$^{*}$ & 8\,MB & 18.9\,MB \\
\bottomrule
\end{tabular}
\caption{\label{tab:scale}Cost of the two released extractions. The EMNLP figure is the amount actually invoiced; the anthology pass ran on donated compute, so its cost is the equivalent at 2026 commercial rates. $^{*}$measured on the deployed instance: the \nClaims{}-claim EMNLP corpus and the 70k-claim six-venue instance built from the anthology extraction.}
\end{table}

\section{Evaluation}
\label{sec:evaluation}
\begin{figure*}[t]
    \centering
    \includegraphics[width=1\linewidth]{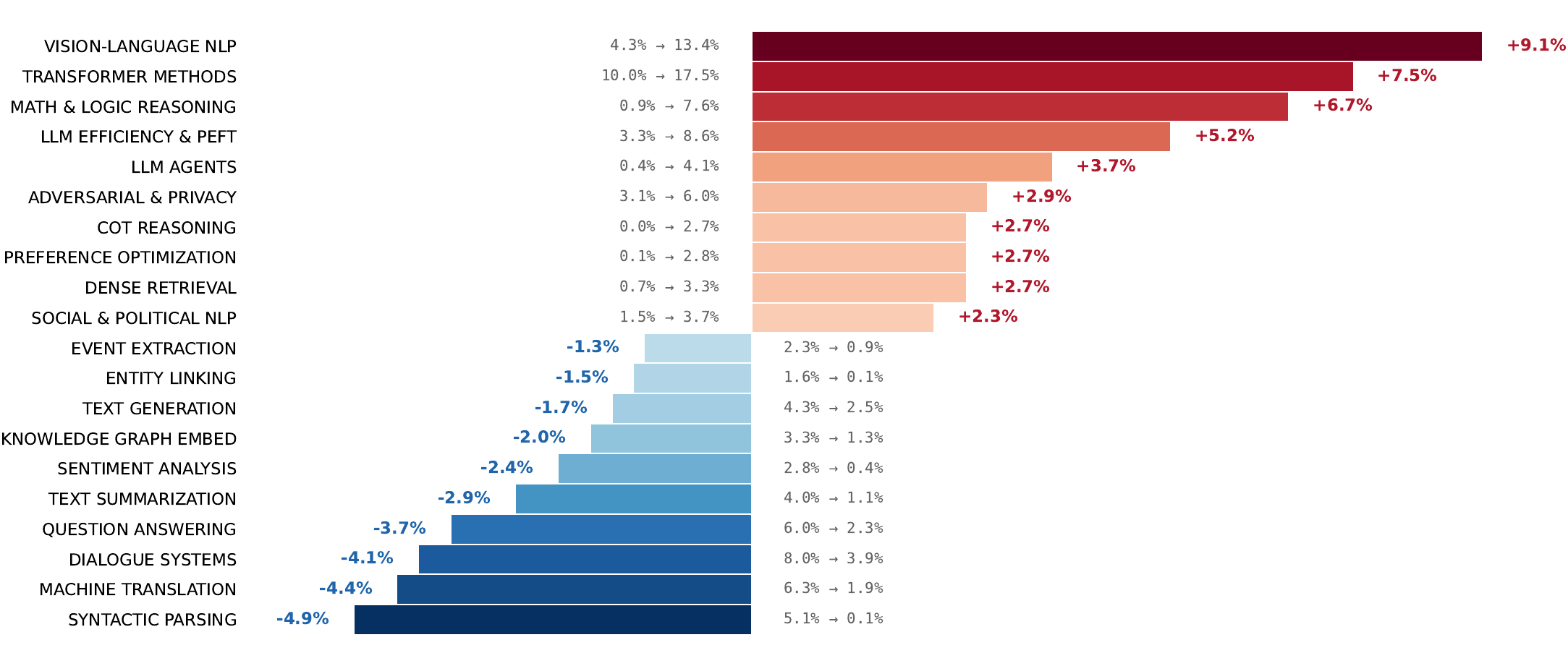}
    \caption{\textbf{Scientific drift in EMNLP (2020--2025)} as measured by the system. Diverging bars show the change in paper-level document frequency (percentage points) for the top emerging and declining ACC clusters, annotated with 2020 and 2025 shares.}
    \label{fig:absolute_shift}
\end{figure*}

We evaluate the two stages that could corrupt the analysis: claim extraction quality and cluster validity. Both use the released EMNLP 2020--2025 ACC corpus: extraction quality on a stratified sample of extracted claims (with curated negative controls), and cluster validity on the 653 papers where our corpus overlaps the SToP gold taxonomy \citep{rohatgi-etal-2023-acl}. Robustness of the headline findings is evaluated separately (\S\ref{subsec:robustness}).

\subsection{Human Validation of Claim Quality}
\label{subsec:human}
Three of the authors labeled 180 items (136 claims sampled from the extraction output, stratified by year, plus 44 curated invalid claims, which the annotators did not know about, as negative controls) as \text{Good}, \text{Bad}, or \text{Unsure}; a claim is \text{Good} if it is atomic, self-contained, contribution-bearing, and faithful to the abstract. Inter-annotator agreement is high and robust to the \text{Unsure} convention (Fleiss' $\kappa = 0.844$ dropping \text{Unsure}, $0.76$ folding it into \text{Bad}, $0.73$ across three categories). On the 136 sampled claims the annotators rate 97.8\%, 91.8\%, and 90.5\% as \text{Good}, and flag 44/44 planted negatives by 2-of-3 majority, so the labels are discriminating, not lenient. An independent LLM judge from a different vendor (Gemini vs.\ Qwen), separately prompted and blind to the human labels, matches the majority consensus with 94.5\% accuracy ($\kappa = 0.863$) and flags 43 of 44 negatives, supporting its use for monitoring extraction on new corpora. The dominant failure modes are unsupported details and background-context extraction; the full protocol is released with the data.

\subsection{Cluster Validity Against an External Taxonomy}
\label{subsec:stop}
We benchmark the clustering against the SToP gold taxonomy \citep{rohatgi-etal-2023-acl} on the 2020--2021 overlap (653 papers), comparing eight configurations: bag-of-words topic models (LDA, NMF), abstract-level neural clustering (BERTopic with SPECTER2 and MPNet encoders), sentence-level clustering (SentSPECTER), and claim-level clustering (ACC, ours), each optionally restricted to its 30 largest clusters. Table~\ref{tab:stop_compact} summarizes the SToP-alignment results.

\begin{table}[t]
\centering
\small
\setlength{\tabcolsep}{1.5pt}
\begin{tabular}{l r r r r r}
\toprule
\textbf{Method} & \textbf{Pur.} & \textbf{LRAP} & \textbf{nDCG@1} & \textbf{V-mes} & \textbf{PairF1} \\
\midrule
LDA & 0.245 & 0.508 & 0.329 & 0.278 & 0.170 \\
NMF & 0.294 & 0.585 & 0.445 & 0.324 & 0.186 \\
BERTopic\textsubscript{SPECTER2} & 0.324 & 0.651 & 0.515 & 0.469 & 0.289 \\
BERTopic\textsubscript{MPNet} & 0.327 & 0.651 & 0.509 & 0.457 & 0.234 \\
SentSPECTER & 0.654 & 0.668 & 0.540 & 0.532 & 0.268 \\
SentSPECTER\textsubscript{top30} & 0.472 & 0.637 & 0.512 & 0.471 & 0.365 \\
ACC \textit{(ours)} & \textbf{0.689} & 0.674 & 0.580 & \textbf{0.571} & 0.361 \\
ACC\textsubscript{top30} & 0.618 & \textbf{0.694} & \textbf{0.596} & 0.552 & \textbf{0.403} \\
\bottomrule
\end{tabular}
\caption{\label{tab:stop_compact}Alignment with the SToP taxonomy (2020--2021 subset, 653 papers). Purity and V-measure are computed on covered documents; LRAP and nDCG@1 via 5$\times$5 repeated cross-validation, the topic-to-label map learned per fold on the training split. The step from abstract-level (BERTopic\textsubscript{SPECTER2}, 0.324) to sentence-level (0.654) to claim-level (0.689) purity shows input granularity matters more than encoder choice. Full table with intrinsic metrics and coverage in the released artifacts.}
\end{table}

Coverage differs by construction: LDA/NMF/BERTopic are scored with a soft document--topic distribution (BERTopic via \texttt{approximate\_distribution}) that assigns mass even to abstracts HDBSCAN treats as outliers (${\sim}$100\% coverage), whereas ACC keeps a hard HDBSCAN partition with noise excluded, leaving all-noise papers uncovered (83.9\%). The coverage-robust V-measure, LRAP, nDCG@1, and top-30 rows neutralize this, and ACC leads or ties throughout. The soft step was required for BERTopic, because under equally hard assignment it leaves 22--26\% of abstracts as outliers vs.\ 16\% for ACC. The hard partition is therefore not what costs coverage: the noise fraction follows from the input granularity. A soft ACC assignment remains a natural extension.

\subsection{Robustness of Headline Findings}
\label{subsec:robustness}
Our headline finding is the set of 15 cluster-level prevalence shifts in Figure~\ref{fig:absolute_shift}. Six reasons could explain the pattern without the field having changed: sampling noise, a two-snapshot artifact, the encoder, the cluster-matching rule, the extractor, and the clustering hyperparameters. Each test below tracks the same quantity, the sign of the 2020$\to$2025 shift.

\textbf{Sampling.} All 15 shifts are significant under a two-proportion $z$-test (13 at $p<0.001$), and their bootstrap 95\% CIs on $\Delta_{DF}$ (2{,}000 paper-level resamples) exclude zero.

\textbf{Two snapshots.} Trajectories are largely monotone across the six years (parsing runs $5.1 \to 3.2 \to 1.6 \to 1.6 \to 0.8 \to 0.1$\% of papers; Figure~\ref{fig:trajectories}), so the shifts are not an artifact of comparing the two endpoints.

\textbf{Encoder.} We re-embed and re-cluster the same claims with nine alternative representations under an identical UMAP/HDBSCAN pipeline: eight other sentence and document encoders (SciNCL, SPECTER1, MPNet, MiniLM, BGE-large/small, E5-large, Nomic) and a non-neural TF-IDF\,+\,SVD baseline. Cluster counts and noise fractions stay in the same range (55--116 clusters, 31--53\% noise), and every representation preserves the sign for at least 11 of the 15 headline clusters (median 14/15; TF-IDF\,+\,SVD 12/15); a cluster in an alternative clustering is identified with a headline cluster by best-Jaccard overlap of their claim sets. Per-encoder results are in Appendix~\ref{sec:app_ablation}.

\textbf{Matching rule.} Recomputing all signs under two sign-blind alternatives to best-Jaccard matching, union overlap and weighted claim image, never lowers a per-encoder count.

\textbf{Extractor.} Re-extracting with \texttt{gpt-oss-120b}, a different model family that yields 4.6 claims per paper against 3.7, preserves all 15 signs and keeps claim-level clustering ahead of every abstract-level baseline on SToP (purity 0.638, V-measure 0.541).

\textbf{Clustering hyperparameters.} A 36-cell sweep (UMAP \textit{n\_neighbors} $\in \{30,40,50\}$, \textit{n\_components} $\in \{5,10\}$; HDBSCAN \textit{min\_cluster\_size} $\in \{20,25,30\}$, \textit{min\_samples} $\in \{5,10\}$) yields clusterings that agree with the reported one at ARI $0.79$--$0.95$ (AMI $0.93$--$0.97$), with between 55 and 116 clusters. The sign is preserved for all 15 headline clusters in all 36 cells.

\section{Case Study: EMNLP 2020--2025}
\label{sec:casestudy}
Running the system on \nProcessedPapers{} EMNLP papers (\nPapers{} per year across 2020--2025) yields the field-level picture in Figure~\ref{fig:absolute_shift}. Classic task clusters lose document-frequency share: parsing falls most steeply ($-4.9$ pp, $-97\%$ relative), while machine translation, dialogue, and question answering each shrink by $3.7$--$4.4$ pp. LLM-era clusters grow in their place: multimodal and vision-language modeling gains $+9.1$ pp ($3.1\times$ its 2020 prevalence) and agents rise more than tenfold ($10.3\times$), with mathematical reasoning, large-model tuning and inference, and a newly emergent LLM reasoning-behavior cluster making up the rest of the growth. Qualitatively, the declining tasks have not vanished: claim search shows their residual claims thinning and dispersing into the noise layer and neighbouring clusters rather than re-forming as standalone contributions --- fragmentation that paper- or keyword-level counts cannot distinguish from disappearance. Figure~\ref{fig:absolute_shift} annotates all twenty clusters with their 2020 and 2025 shares.

\subsection{Use Cases}
\label{subsec:usecases}
\textbf{Is this area still growing?} A researcher weighing a retrieval project wants to know whether the area is expanding before committing to it. Typing \texttt{retrieval} into the Map search highlights every claim whose text or source title matches, and ranks clusters by how many of those matches they hold. The top-ranked cluster's trend page shows its share of papers rising from 0.7\% in 2020 to 3.3\% in 2025, and the ranking places that growth in dense-retrieval and LLM-adjacent clusters rather than in classic IR. Each point on the trajectory opens the claim behind it and links to its paper.

\begin{figure}[t]
\centering
\includegraphics[width=\linewidth]{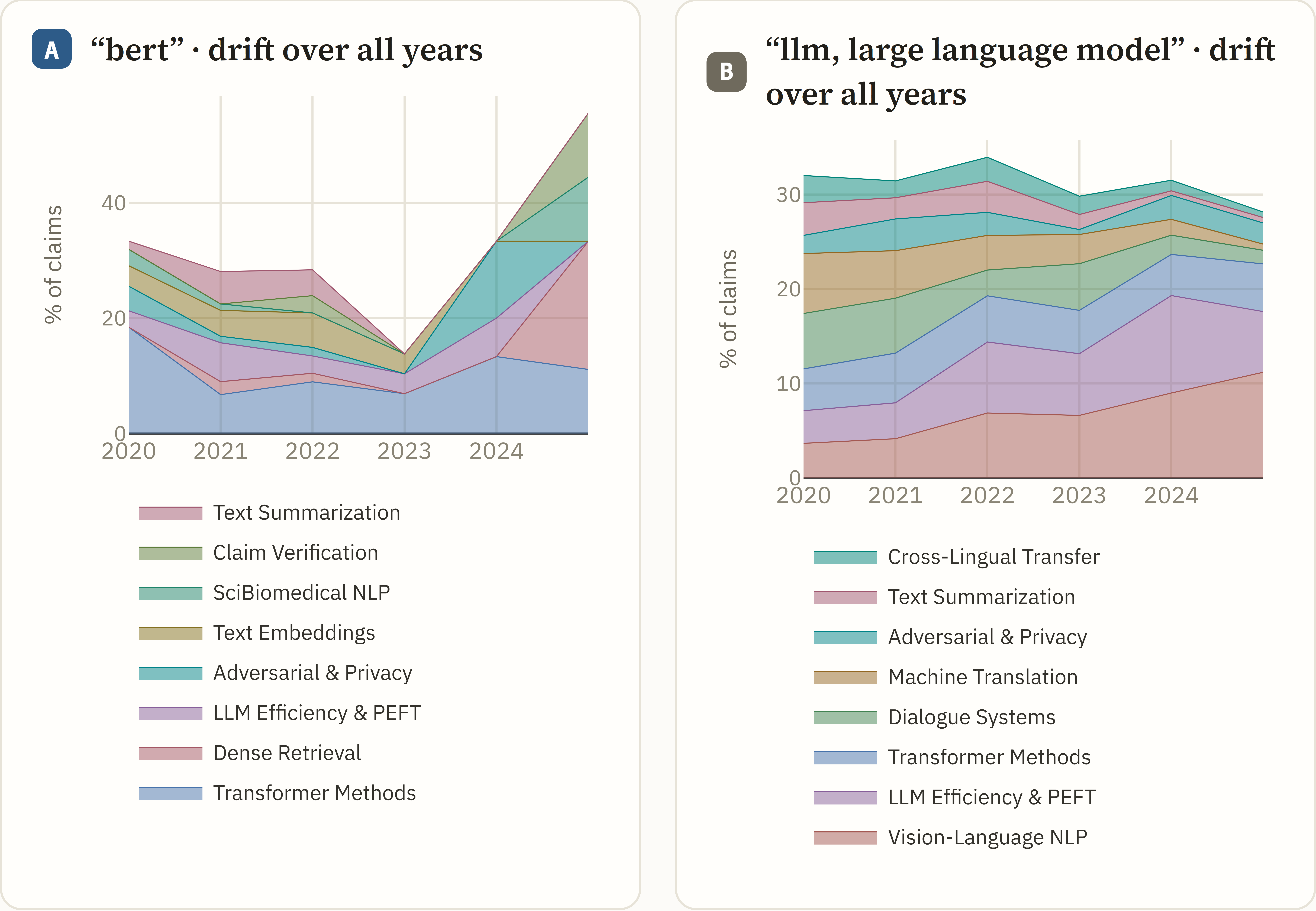}
\vspace{-4mm}
\caption{\textbf{Compare view, \emph{over time} mode:} yearly topic mix (top clusters, \% of cohort claims) for the \texttt{bert} cohort (197 papers) vs.\ \texttt{llm} / \texttt{large language model} (3{,}253 papers): the residual \texttt{bert} agenda concentrates in claim verification and biomedical NLP, while the \texttt{llm} cohort shifts toward vision--language modeling and efficiency.}
\label{fig:compare}
\vspace{-3mm}
\end{figure}

\textbf{What does a venue publish?} Compare takes two cohorts and ranks clusters by their gap in paper share. Figure~\ref{fig:aclcoling} contrasts ACL with COLING: COLING keeps a stronger classic-NLP profile (embeddings, aspect-based sentiment, NER, event extraction), while ACL leans toward efficient reasoning, multilingual LLMs and vision--language evaluation. A cohort can also be an author or a keyword-defined subsample, so \texttt{bert} against \texttt{llm} (Figure~\ref{fig:compare}) or two research groups. The \emph{over time} mode makes the comparison diachronic.

\textbf{What is emerging, and what is declining?} Trends ranks every cluster by its 2020$\to$2025 shift with bootstrap intervals, so both ends appear in one view: LLM reasoning behaviour and vision--language modeling at the growing end, syntactic parsing ($-4.9$ pp, $-97\%$ relative) at the other. Because prevalence counts papers with at least one claim in a cluster, a topic that is still discussed but no longer contributed looks different from one that has disappeared: claim search shows the residual parsing claims scattered across the noise layer and neighbouring clusters instead of forming a cluster of their own. A narrower term behaves the same way. Searching \texttt{humor} puts almost all of its matches in the noise layer, each still linked to its source paper.

\section{Conclusion}
Drift Inspector turns a validated claim-extraction pipeline into an interactive instrument for watching a field reorganize itself, its trends reflecting what papers assert they add rather than the vocabulary of their motivation. The system is fully open (MIT code, CC~BY data) and corpus-agnostic: any venue with abstracts can be mapped in one pass. Next step is implementing a hierarchical theme system aggregating clusters into navigable super-themes.

\section*{Limitations}
The analysis relies on abstracts, which omit technical nuance and negative results, and some contributions surface only in the paper body. All validated results are on a single venue (EMNLP): the larger extractions we release (six *ACL venues; the full anthology) are only LLM-judge-monitored, not human-revalidated, so their per-cluster numbers warrant more caution. The pipeline rests on an LLM extractor that can introduce errors (hallucination, under/over-splitting), mitigated but not eliminated by schema and human validation and an aligned LLM judge; human validation was by the authors on 136 claims, so agreement partly reflects shared training, and the negatives are curated corruptions rather than sampled natural errors. Clustering depends on UMAP/HDBSCAN hyperparameters and the embedding model; headline drift directions are stable under bootstrap resampling, eight encoders, and a lexical representation, but cluster \emph{boundaries} and labels may vary, and drift statistics are descriptive, not causal.

\section*{Ethics Statement}
The system analyzes publicly available ACL Anthology abstracts (CC~BY~4.0) and involves no private data. The main risk is misinterpretation: cluster prevalence reflects sampling, extraction, and clustering choices and should be read as a descriptive signal about abstracts, not a verdict on subfields' scientific value: a decline in standalone parsing contributions does not mean parsing is solved or worthless. We document all parameters and release all artifacts to keep the measurement inspectable.

\bibliography{custom}
\clearpage
\appendix

\begin{figure*}[t]
\centering
\includegraphics[width=\textwidth]{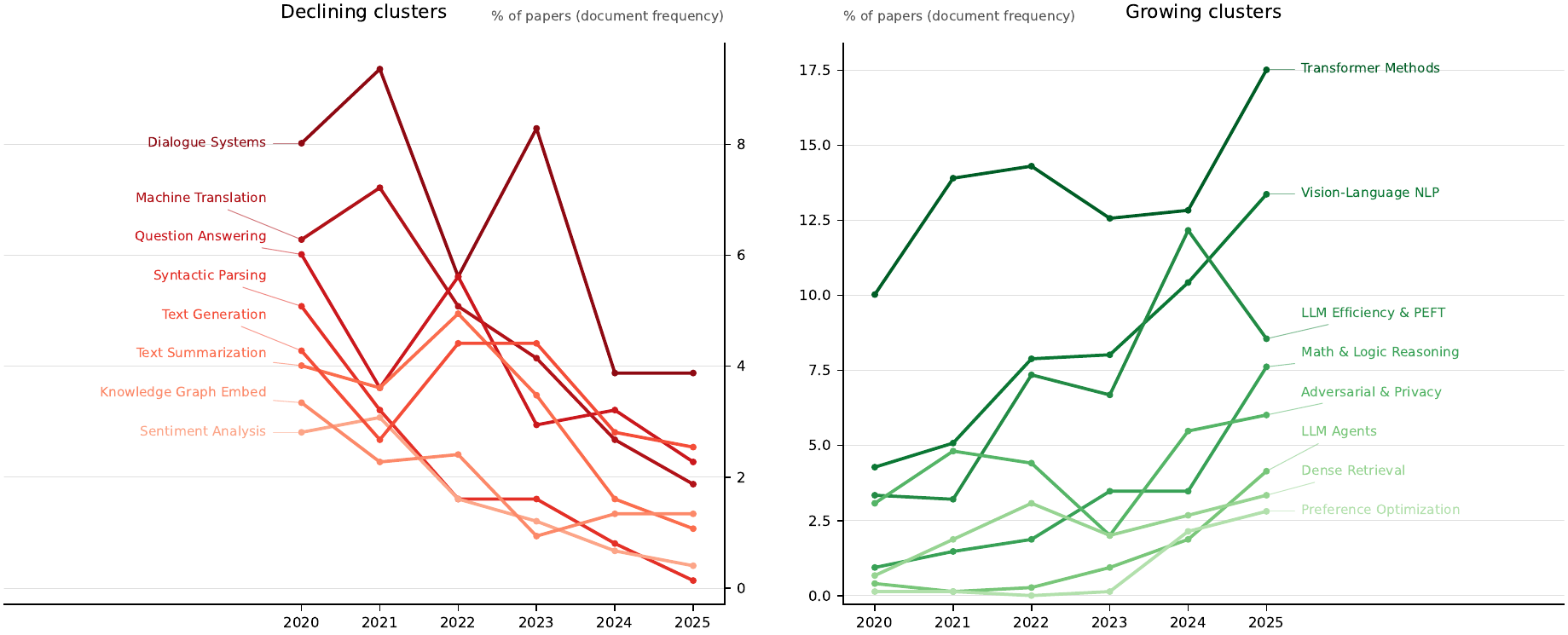}
\caption{Per-year document-frequency trajectories for the eight most declining and eight most growing ACC clusters (paper-level DF, \% of papers), computed from the canonical clustering. Labels are curated short cluster names; full names and c-TF-IDF descriptors ship with the released data. Headline endpoint shifts are significant with bootstrap CIs excluding zero (\S\ref{subsec:robustness}).}
\label{fig:trajectories}
\end{figure*}

\section{Extraction and Judge Prompts (Condensed)}
\label{sec:app_extraction}

The extractor returns JSON \texttt{\{"claims": [\{"text": ...\}]\}}. Core instruction (few-shot examples omitted; full prompts in the repository):

\begin{quote}\small
\textbf{Task.} Extract Atomic Contribution Claims (ACCs) from NLP paper titles and abstracts --- faithful, self-contained, atomic, contribution-bearing propositions about what the paper introduces, proposes, evaluates, demonstrates, or establishes.

\textbf{Include:} new methods, models, objectives, training/inference procedures; new datasets, benchmarks, resources, tools, evaluation setups; empirical findings and analyses established by the paper. \textbf{Exclude:} background, motivation, related work, future work, release logistics; vague problem statements; prior-work claims; raw metric-only claims.

\textbf{Rules:} one proposition per claim; no facts absent from the title/abstract; resolve pronouns and vague references; no meta-language (``this paper'', ``we''); mid-level granularity; no raw numbers; no near-duplicates; if unsure, omit; usually 1--5 claims, empty list if none.

\textbf{Self-validation.} Silently check every candidate against: contribution-bearing, atomic, faithful, decontextualized; discard failures.
\end{quote}

The independent LLM judge receives one candidate claim with its title and abstract and labels it \text{Good}/\text{Bad}/\text{Unsure} with a main-issue category (\texttt{background\_context}, \texttt{unsupported}, \texttt{too\_vague}, \texttt{not\_self\_contained}, \texttt{not\_contribution}, \texttt{overgeneralized}, \texttt{mixed\_claims}); it is prompted separately from the extractor and never sees human labels.

\textbf{Human-validation sampling.} We stratify the 180-item sample by year (seed 42; 54/18/18/18/18/54 across 2020--2025). Of these, 136 are claims from the \nClaims{}-claim output, at most one per paper, and 44 are negative controls: invalid claims we built by injecting an unsupported detail, recasting motivation as a contribution, or over-generalizing. Annotators see only the title, abstract, and claim, and never learn which items are controls.

\section{Pipeline Configuration}
\label{sec:app_representation}
\textbf{Corpus.} EMNLP main-track 2020--2025 from the ACL Anthology; each year balanced to \nPapers{} non-empty papers (the smallest usable year, 2020, has 751 papers of which 3 have no extractable contribution), seed 42, \nProcessedPapers{} papers and \nClaims{} claims in the analysis corpus. We do not balance the scale corpora by year: the six-venue instance keeps every main-track paper of its six venues (69{,}950 claims, 88 clusters, 2018--2026).

\textbf{Extractor.} \texttt{\extractor} via OpenRouter, temperature 0.2, top-$p$ 0.9, JSON-object mode, no \textit{max\_tokens} cap; unparsable outputs dropped. It covered 4{,}960 abstracts (\nProcessedPapers{} after year-balancing) for 42.1M billed tokens, 82\% of them reasoning the caller never sees. The anthology used \texttt{gpt-oss-120b}, \textit{reasoning\_effort} medium, strict JSON schema, \textit{max\_tokens} 8000, temperature 0.

\textbf{Embedding.} \texttt{allenai/}\allowbreak\texttt{specter2\_aug2023refresh\_base} with the proximity adapter; claim text only, CLS pooling, max length 512.

\textbf{Clustering.} UMAP: \textit{n\_neighbors}=40, \textit{n\_components}=5 (clustering) and 2 (visualization), \textit{metric}=cosine, seed 42. HDBSCAN: \textit{min\_cluster\_size}=25, \textit{min\_samples}=5, \textit{eom} selection; noise label $-1$. Descriptors: class-based TF-IDF \citep{Grootendorst2022BERTopicNT} over aggregated cluster claims with hyphen-aware (1,2)-gram tokenisation, a claim-frequency floor ($\geq$10 claims), a per-cluster coverage filter ($\geq$5\%), and a Snowball-stemmer de-duplicator collapsing morphological/spelling variants; short (top-3) and extended (top-5) descriptors. All 80 clusters carry LLM-generated, author-reviewed short/full names; tables and figures use the short name, falling back to the descriptor.

\begin{figure}[t]
\centering
\includegraphics[width=\linewidth]{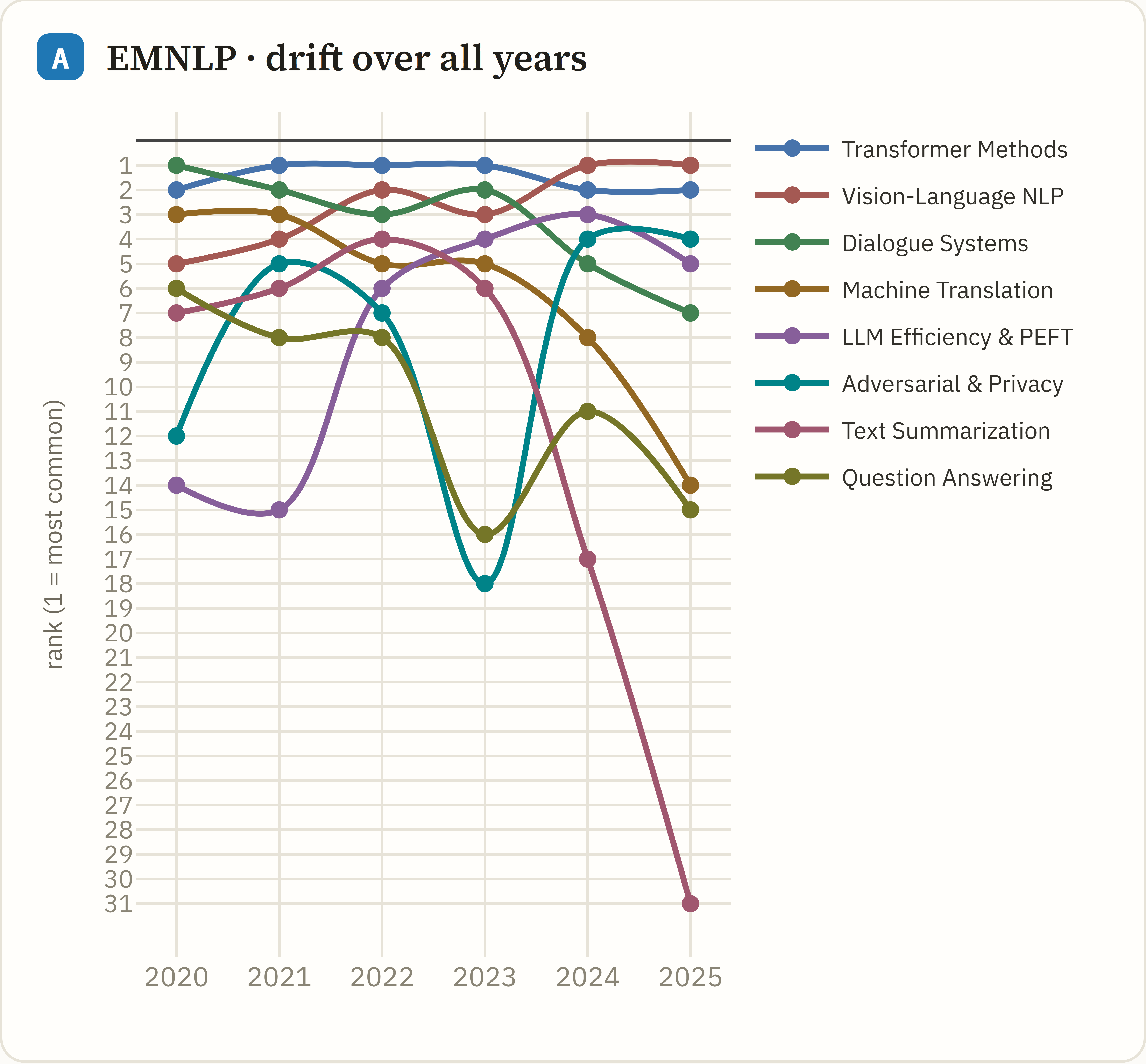}
\vspace{-4mm}
\caption{Single-conference diachronic view (Compare, \emph{over time}, bump chart): EMNLP's top clusters ranked by yearly document frequency (1 = most common). Vision--language modeling reaches rank~1 by 2024; LLM efficiency climbs from 14 to 3--5; summarization falls from 7 to 31.}
\label{fig:bump}
\end{figure}

\textbf{Frontend.} Year, color, search, and theme changes re-render through \texttt{Plotly.react}; search and click-through run client-side over per-point metadata. The portable single-file build embeds css, js, Plotly, and data ($\sim$8~MB for EMNLP, 18.9~MB at six-venue scale), with web fonts the only external request. Figure~\ref{fig:bump} shows Compare's bump-chart mode.

\section{Encoder Ablation}
\label{sec:app_ablation}

Table~\ref{tab:encoder_ablation} re-embeds and re-clusters the \nClaims{} claims with nine alternative representations under the identical UMAP/HDBSCAN pipeline (\S\ref{subsec:robustness}). ARI is measured against the canonical SPECTER2 clustering and SToP purity on the 653-paper overlap of Table~\ref{tab:stop_compact}; signs are matched by best Jaccard, which is sign-blind by construction. Recomputing all $9\times15$ signs under two alternative criteria never lowers a per-encoder count. Every representation preserves at least 11/15 signs, so the drift is not an artifact of the encoder, and SPECTER2's top purity is why we make it canonical.

\begin{table}[t]
\centering
\small
\setlength{\tabcolsep}{3pt}
\begin{tabular}{l r r r r r}
\toprule
\textbf{Encoder} & \textbf{$k$} & \textbf{Noise\%} & \textbf{ARI} & \textbf{SToP} & \textbf{Sign} \\
\midrule
SPECTER2 \textit{(ours)} & 80 & 36.1 & 1.00 & \textbf{0.689} & 15/15 \\
SciNCL & 111 & 31.2 & 0.75 & 0.670 & 15/15 \\
SPECTER1 & 85 & 34.1 & 0.69 & 0.658 & 14/15 \\
MPNet & 116 & 32.6 & 0.61 & 0.681 & 15/15 \\
MiniLM & 107 & 37.1 & 0.64 & 0.646 & 14/15 \\
BGE-large & 99 & 37.9 & 0.67 & 0.674 & 13/15 \\
BGE-small & 92 & 39.0 & 0.64 & 0.657 & 14/15 \\
Nomic & 108 & 40.6 & 0.50 & 0.670 & 14/15 \\
E5-large & 55 & 52.7 & 0.37 & 0.577 & 11/15 \\
TF-IDF\,+\,SVD & --- & --- & --- & --- & 12/15 \\
\bottomrule
\end{tabular}
\caption{\label{tab:encoder_ablation}Encoder ablation on the \nClaims{} claims. \textbf{$k$}: clusters, noise excluded. \textbf{Sign}: headline shift signs preserved of 15. TF-IDF\,+\,SVD is a non-neural control, scored on sign preservation only.}
\end{table}

\end{document}